%% file: main.tex
\documentclass{article}

\usepackage{microtype}
\usepackage{graphicx}
\usepackage{booktabs} 
\usepackage{hyperref}

\usepackage[preprint]{icml2026}

\usepackage{amsmath}
\usepackage{amssymb}
\usepackage{mathtools}
\usepackage{amsthm}

\usepackage[capitalize,noabbrev]{cleveref}

\theoremstyle{plain}

\theoremstyle{definition}

\theoremstyle{remark}

\usepackage[textsize=tiny]{todonotes}

\usepackage{times}
\usepackage{soul}
\usepackage{url}
\usepackage{caption}
\usepackage{amsmath}
\usepackage{amsthm}
\usepackage{booktabs}
\usepackage{algorithm}
\usepackage{algorithmic}
\usepackage[switch]{lineno}

\usepackage{textcomp}
\usepackage{xcolor}
\usepackage{amssymb}
\usepackage{threeparttable}
\usepackage{booktabs}
\def\BibTeX{{\rm B\kern-.05em{\sc i\kern-.025em b}\kern-.08em
    T\kern-.1667em\lower.7ex\hbox{E}\kern-.125emX}}

\usepackage{amsfonts}
\usepackage{multirow}

\usepackage{tikz}
\newcommand*\emptycirc[1][.75ex]{\tikz\draw (0,0) circle (#1);} 
\newcommand*\fullcirc[1][.75ex]{\tikz\fill (0,0) circle (#1);} 

\usepackage{enumitem}
\usepackage{subcaption}
\usepackage{comment}
\usepackage{xspace}
\usepackage{hyperref}
\usepackage{cleveref}

\usepackage{pifont}

\definecolor{DeepGreen}{RGB}{0,120,0}
\definecolor{DeepRed}{RGB}{180,0,0}
\definecolor{Deepblue}{RGB}{0, 0, 146}

\newcommand{\ours}{\textsc{FairT2V}\xspace}

\newif\ifdraft
\draftfalse   

\ifdraft
  
  \newcommand{\rev}[1]{\textcolor{red}{#1}}
    \newcommand{\ym}[1]{\textcolor{blue}{#1}}
     \newcommand{\morri}[1]{\textcolor{cyan}{#1}}
\else
  
  \newcommand{\rev}[1]{#1}
    \newcommand{\ym}[1]{#1}
     \newcommand{\morri}[1]{#1}
\fi

\newcommand{\neu}{\text{neu}\xspace}
\newcommand{\maj}{\text{maj}\xspace}

\newcommand{\calT}{\mathcal{T}}
\newcommand{\NeuG}{\calT_{\neu}}
\newcommand{\MajG}{\calT_{\maj}}
\newcommand{\MinG}{\calT_{\min}}

\usepackage{makecell}

\usepackage[table]{xcolor}   
\usepackage{colortbl}

\usepackage{tcolorbox}
\newtcolorbox{prompt}[1]{
    colback=gray!5!white,
    colframe=gray!40!black,
    coltitle=black,
    colbacktitle=gray!35!white,
    fonttitle=\bfseries,
    title=#1,
    boxrule=.4pt,
    left=1mm,
    right=1mm,
    top=1mm,
    bottom=1mm
}

\icmltitlerunning{FAIRT2V: Training-Free Debiasing for Text-to-Video Diffusion Models}

\begin{document}

\twocolumn[
\icmltitle{FAIRT2V: Training-Free Debiasing for Text-to-Video Diffusion Models}



\icmlsetsymbol{equal}{*}

\begin{icmlauthorlist}
\icmlauthor{Haonan Zhong}{unsw}
\icmlauthor{Wei Song$^\dagger$}{unsw}
\icmlauthor{Tingxu Han}{nju}
\icmlauthor{Maurice Pagnucco}{unsw}
\icmlauthor{Jingling Xue}{unsw}
\icmlauthor{Yang Song}{unsw}

\end{icmlauthorlist}

\icmlaffiliation{unsw}{School of Computer Science and Engineering, University of New South Wales, Australia.}
\icmlaffiliation{nju}{State Key Laboratory for Novel Software Technology, Nanjing University, China}

\icmlcorrespondingauthor{Wei Song}{wei.song1@unsw.edu.au}

\icmlkeywords{Machine Learning, ICML}

\vskip 0.3in
]



\printAffiliationsAndNotice{}  

\input{tex/abstract}
\input{tex/introduction}
\input{tex/related_work_2}
\input{tex/method}
\input{tex/experiment}
\input{tex/conclusion}

\section*{Impact Statement}
This paper aims to analyze and mitigate demographic bias in text-to-video diffusion models by introducing \ours, a training-free framework that adjusts text-conditioning embeddings during inference. We expect \ours to help researchers and practitioners better understand how demographic bias propagates in T2V systems, assess fairness–quality trade-offs more systematically, and develop fairer video generation systems.

Our work also reveals potential risks associated with bias propagation in generative video models. We show that implicit demographic associations can persist across frames and reinforce stereotypical representations even under neutral prompts, which may negatively affect downstream applications. The proposed debiasing procedure is introduced to demonstrate how such bias can be mitigated in a lightweight and controllable manner, motivating the development of more fairness-aware T2V models. We do not advocate the use of embedding-level interventions to override explicit user intent or to deploy fairness guarantees without domain-specific consideration.

All experiments are conducted solely for research and evaluation purposes in controlled settings. We encourage responsible use of these techniques, including clear documentation of their scope, transparent reporting of evaluation protocols, and careful consideration of deployment contexts. Overall, we believe the net societal impact of this work is positive, as improved analysis and mitigation of demographic bias in T2V generation may reduce unintended representational harms and support fairer video generation systems.


\bibliography{icml}
\bibliographystyle{icml2026}

\newpage
\appendix
\onecolumn
\input{appendix}



\end{document}

%% file: tex/abstract.tex
\begin{abstract}
\rev{Text-to-video (T2V) diffusion models have achieved rapid progress, yet their demographic biases—particularly gender bias—remain largely unexplored.} We present \ours, \rev{a training-free debiasing framework for text-to-video generation that mitigates encoder-induced bias without fine-tuning.} We first analyze demographic bias in T2V models and show that it primarily originates from pretrained text encoders, which encode implicit gender associations even for neutral prompts. \rev{We quantify this effect with a gender-leaning score that correlates with bias in generated videos.}

Based on this insight, \ours mitigates demographic bias by neutralizing prompt embeddings via anchor-based \rev{spherical geodesic transformations} while preserving semantics. To maintain temporal coherence, \rev{we apply debiasing only during early identity-forming steps through a dynamic denoising schedule.} We further propose a video-level fairness evaluation protocol combining \rev{VideoLLM-based reasoning} with human verification. Experiments on the modern T2V model Open-Sora show that \ours substantially reduces demographic bias \rev{across occupations} with minimal impact on video quality.
\end{abstract}

%% file: tex/introduction.tex
\section{Introduction}
\label{sec:intro}

Text-to-video (T2V) models have progressed rapidly in recent years, driven by advances in diffusion-based architectures~\cite{DiT, DALLE, sora, CogVideoX, Open-Sora}. These systems are increasingly deployed in applications such as advertising, education, and entertainment. However, recent studies~\cite{sd-bias-b1, sd-bias-b2, sd-bias-b3, sora-bias} show that modern generative models frequently over-represent gender, racial, or age groups. Such demographic biases raise significant ethical concerns, as they risk reinforcing harmful societal stereotypes and perpetuating representational inequities~\cite{Fair-Diffusion, FairT2I}.

Despite growing interest in fairness for generative models, demographic bias in text-to-video generation remains largely unexplored. Existing studies have focused primarily on text-to-image (T2I) generation, leaving open the question of how bias manifests and propagates in video generation, where temporal dynamics, identity persistence, and repeated text conditioning introduce new challenges. 
\ym{In this work, we focus on gender bias associated with occupation, as it is a well-documented and measurable form of demographic bias that enables controlled analysis in the text-to-video setting.}

\rev{Unlike images, gender-related bias in videos can be reinforced across frames, leading to temporally persistent stereotypes. Moreover, long-range temporal structure and multiple identities make frame-wise or image-level fairness interventions insufficient and inconsistent.}

In this paper, we present \ours, the first systematic study of demographic bias in text-to-video generation. We formalize implicit gender associations in neutral prompt embeddings via a gender-leaning score, enabling quantitative analysis of bias at the text-conditioning level. Using this formulation, we show that demographic bias in T2V models primarily originates from pretrained text encoders, which encode gender associations even without explicit demographic cues. This bias arises because encoders such as CLIP, trained on large-scale but socially imbalanced image--text data and widely used in modern T2V architectures, internalize skewed correlations that disproportionately align certain occupations with gender-associated directions in the prompt embedding space.

Based on this insight, we introduce \ours, a training-free debiasing framework for text-to-video generation. \ours operates in an angular embedding space, where prompt embeddings and demographic attribute anchors lie on a shared hyperspherical manifold. By adaptively shifting each prompt embedding toward a neutral point based on its relative angular proximity to majority and minority anchors, \ours mitigates encoder-induced bias while preserving prompt semantics, without requiring any model fine-tuning. While this embedding-level debiasing is already effective, directly applying the debiased embeddings throughout the entire diffusion process can disrupt temporal coherence and introduce frame-level artifacts. To further improve video quality, \ours employs a dynamic denoising schedule that applies debiasing only during early identity-forming stages and restores the original prompt embedding during later refinement steps, preserving temporal consistency and visual smoothness~\cite{denoise-effect}.

Finally, we design a fairness evaluation protocol tailored to text-to-video generation. Existing image-based fairness assessments do not readily extend to videos, where multiple identities and dynamic temporal visibility complicate demographic attribution. To address this, we employ a VideoLLM (Gemini~\cite{gemini25flash2025}) to analyze full video sequences and infer demographic attributes by reasoning over identity continuity, temporal transitions, and multi-subject scenes. Since automated VideoLLM-based judgments may be affected by hallucination and prompt sensitivity, we further incorporate a human-in-the-loop verification stage to validate demographic labels and assess video quality.

In summary, this paper makes the following contributions:
\begin{itemize}[leftmargin=*, topsep=0pt, itemsep=0pt]
\item We propose \ours, a training-free debiasing framework for text-to-video generation that mitigates text-encoder--induced demographic bias while preserving prompt semantics and temporal coherence in videos.

\item We provide the first systematic analysis of demographic bias in text-to-video diffusion models, and identify the text-conditioning encoder as a primary source through which demographic bias is introduced and propagated.

\item We develop a video-oriented fairness evaluation protocol that combines LLM-as-a-Judge with human-in-the-loop verification, enabling reliable demographic bias assessment beyond frame-level analysis.
\item We conduct extensive experiments on state-of-the-art text-to-video model, demonstrating that \ours substantially reduces demographic bias while maintaining high video quality across multiple evaluation metrics.
\end{itemize}

%% file: tex/related_work_2.tex
\section{Background and Related Work}
\label{sec:related_work}

\subsection{Text-to-Video Diffusion Models}

Text-to-video (T2V) models such as CogVideoX and Open-Sora extend diffusion transformers to the video domain via MM-DiT architectures. Although effective in practice, MM-DiT--based models remain less well understood than U-Net--based diffusion models, and their spatiotemporal attention behavior under textual conditioning has received limited study. Despite architectural differences, T2V models follow a standard text-conditioned diffusion paradigm. Video generation starts from a noisy latent sequence and proceeds through iterative denoising, guided at each step by embeddings produced by pretrained text encoders such as CLIP, T5, or their combination. The text embedding acts as a persistent conditioning signal throughout the reverse diffusion process, shaping both global structure and identity-level semantics across frames.

Formally, given a text embedding $E_p$ for prompt $p$ and an unconditional embedding $E_{\varnothing}$, classifier-free guidance (CFG) computes the guided noise estimate at step $t$ as:
\begin{equation}
\begin{aligned}
\hat{\epsilon}_\theta(x_t, t, E_p)
& =
\epsilon_\theta(x_t, t,E_{\varnothing}) \\
& \quad
+ \alpha \big(
    \epsilon_\theta(x_t, t, E_p)
    - \epsilon_\theta(x_t, t, E_{\varnothing})
  \big)
\label{eq:cfg}
\end{aligned}
\end{equation}
where $\epsilon_\theta$ denotes the denoising network, $x_t$ is the latent at step $t$, and $\alpha$ is the guidance scale. The latent is updated as:
\begin{equation}
x_{t+1} = \mathcal{D}\big(x_t, t, \hat{\epsilon}_\theta(x_t, t, E_p)\big)
\label{eq:ddpm-update}
\end{equation}
Under classifier-free guidance, each denoising step is a deterministic function of the text embedding $E_p$.

\begin{figure}[t]
    \centering
    \includegraphics[height=3.5cm]{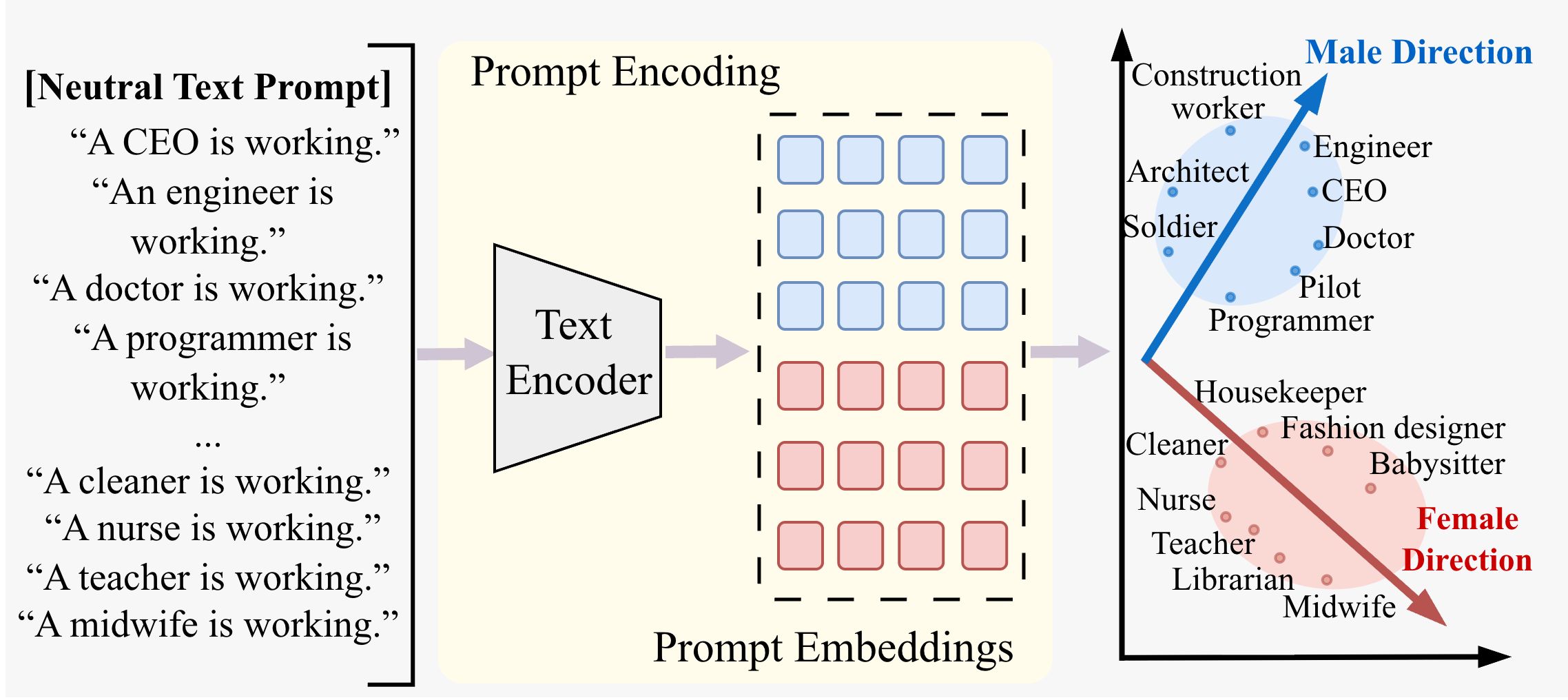}
    \caption{Bias source analysis in text-to-video generation. Neutral prompts are encoded by the text encoder (e.g., CLIP) into embeddings aligned with gender-associated directions, revealing implicit demographic bias in the text-conditioning space.}
    \label{fig:fig4a}
    \vspace{-2ex}
\end{figure}

\subsection{Demographic Bias in Generative Models}

Diffusion-based generative models are known to produce biased or stereotypical outputs even under neutral prompts~\cite{debiasdiff}. Prior analyses of Stable Diffusion~\cite{stable-diff, sd-bias-b1, sd-bias-b2, sd-bias-b3} reveal systematic associations between occupations, visual attributes, and demographic groups, indicating that bias can arise implicitly from text conditioning alone. While most existing studies focus on text-to-image generation, emerging evidence suggests that such biases carry over to video generation. A recent study of Sora~\cite{sora-bias}, for example, reports pronounced gender associations in text-to-video outputs, where particular occupations are consistently linked to one gender despite neutral prompts. These observations motivate the need for a systematic investigation of demographic bias in text-to-video diffusion models.

\begin{figure*}[!t]
    \centering

    \begin{minipage}[t]{0.48\linewidth}
        \centering
        \includegraphics[width=\linewidth]{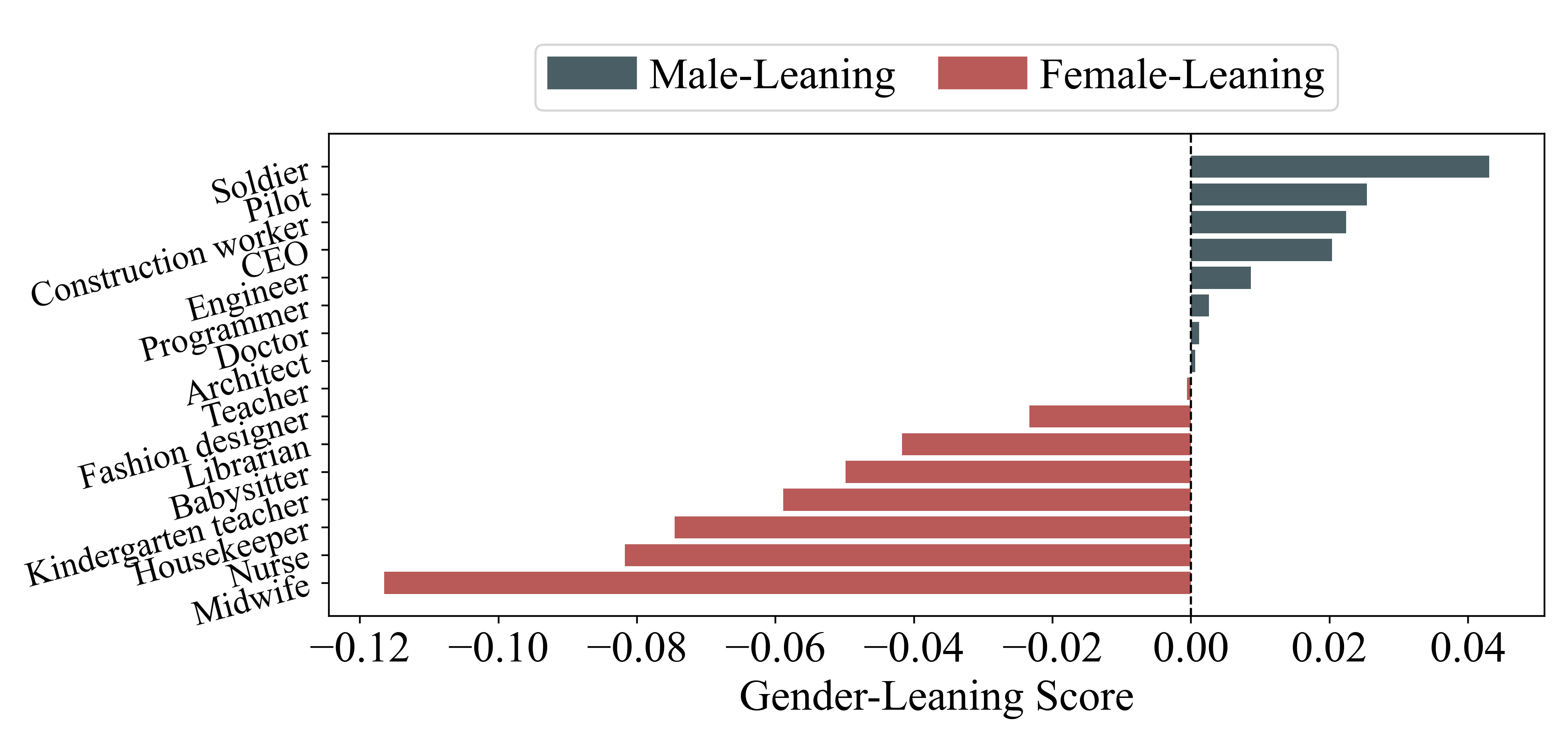}
        \caption{Gender-leaning scores (\Cref{eq:score}) from the CLIP text encoder for 16 occupations, using the prompt sets in \Cref{eq:prompt_sets}.
        }
        \label{fig:fig2}
    \end{minipage}
    \hfill
    \begin{minipage}[t]{0.48\linewidth}
        \centering
        \includegraphics[width=\linewidth]{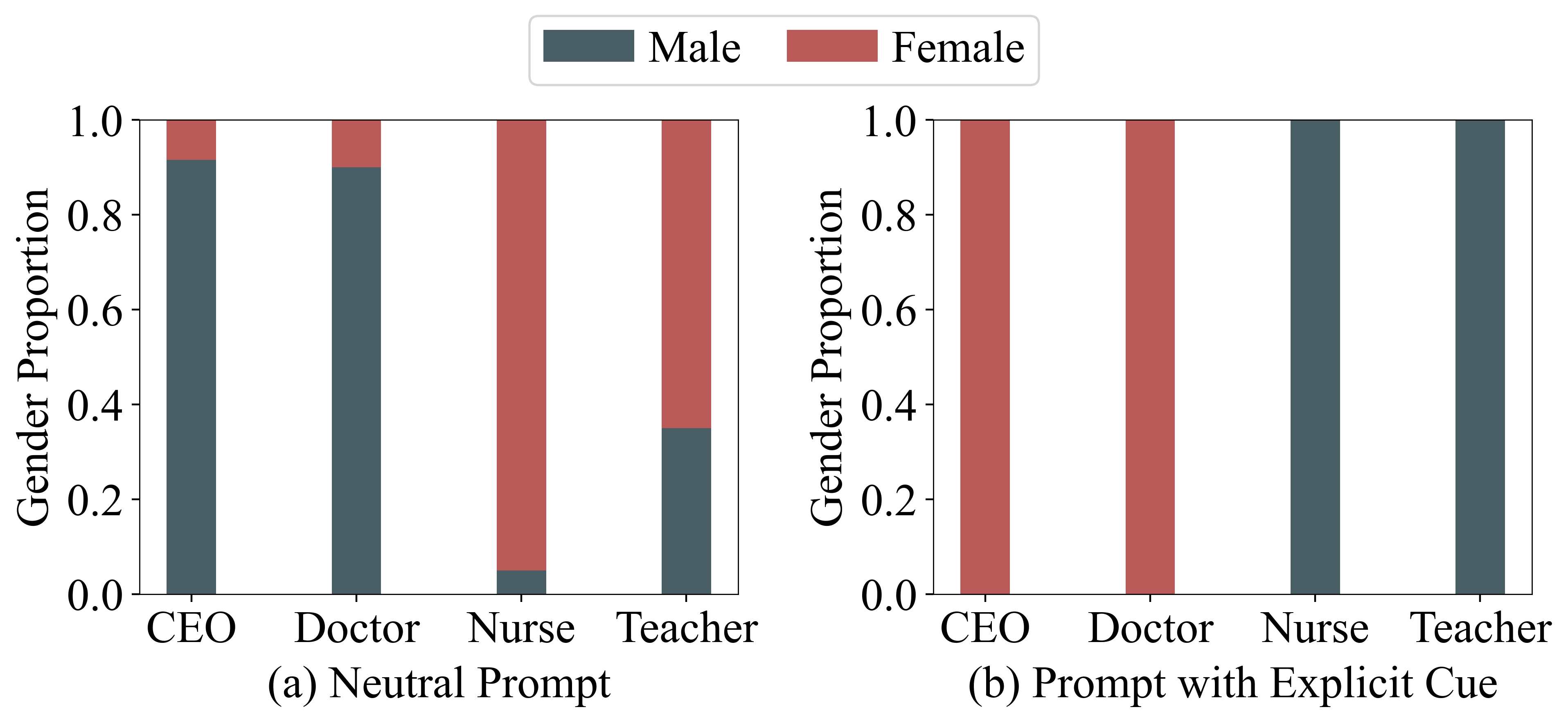}
        \caption{Gender proportions in generated videos for four occupations using neutral prompts (left) and explicit gender cues (right).
        }
        \label{fig:fig3}
    \end{minipage}
\vspace{-2ex}
\end{figure*}

Existing debiasing approaches for diffusion models focus almost exclusively on text-to-image (T2I) generation~\cite{ENTIGEN, Fair-Diffusion, FairT2I, SDID, TBIE, FairImagen, ITI-GEN, unet-h-space, DiffLens, FairQueue}, and can be broadly categorized into training-based and training-free methods. Training-based approaches mitigate bias by fine-tuning generative models or suppressing biased latent directions~\cite{ITI-GEN, unet-h-space, DiffLens, debiasdiff}, but incur substantial computational cost due to additional training, limiting their scalability beyond image generation~\cite{ENTIGEN, Fair-Diffusion, FairT2I}. In contrast, training-free methods mitigate bias by modifying prompts or adjusting text embeddings~\cite{ENTIGEN, FairT2I, Fair-Diffusion, FairImagen}. While computationally efficient, these approaches often introduce semantic drift that alters the original prompt intent and are designed for static image generation, without accounting for the temporal consistency and identity persistence required in video generation.

%% file: tex/method.tex
\section{\ours: \ym{Gender Bias Analysis}}
\label{sec:bias-source}

In text-to-video diffusion models, classifier-free guidance injects the same text embedding at every denoising step, \ym{causing any gender bias in the text encoder to be repeatedly reinforced across frames}. Motivated by this observation, we analyze bias in the text-conditioning pathway. As shown in \Cref{fig:fig4a}, even under gender-neutral prompts, the text encoder maps occupation descriptions to embeddings that form clear gender-correlated clusters, revealing implicit gender associations in the prompt embedding space.

To quantify this effect, we analyze how a pretrained text encoder (CLIP) encodes 16 common occupations (U.S. Bureau of Labor Statistics), 
\morri{which exhibit well-documented gender imbalances in real-world employment}, 
$O = \{\text{CEO, Doctor, Nurse, Teacher, \ldots}\}$, using three prompt sets: gender-neutral, majority-group, and minority-group. For each occupation $o_i$, we define:
\begin{equation}
\small
\label{eq:prompt_sets}
\begin{aligned}
\NeuG(o_i) &= \{\texttt{A/An } o_i \; d_j \mid d_j \in D\} \\
\MajG(o_i) &= \{\texttt{A male } o_i \; d_j \mid d_j \in D\} \\
\MinG(o_i) &= \{\texttt{A female } o_i \; d_j \mid d_j \in D\}
\end{aligned}
\end{equation}
where $D$ is a set of activity modifiers appended to a fixed prompt template
(e.g., \textit{is working in an office}, \textit{is writing a report}, \ldots).
Each prompt set is encoded by the text encoder $\phi(\cdot)$, yielding embeddings
$\neu_{o_i}$, $\maj_{o_i}$, and $\min_{o_i}$.

We measure occupation-specific bias using a local bias index
$\mathrm{BI}_{o_i}=\langle \neu_{o_i},\maj_{o_i}\rangle-\langle \neu_{o_i},\min_{o_i}\rangle$,
which indicates whether the neutral embedding is closer to the majority or minority group. The corresponding local gender direction is defined as
$l_{o_i}=\frac{\maj_{o_i}-\min_{o_i}}{\|\maj_{o_i}-\min_{o_i}\|_2}$,
with its sign aligned to $\mathrm{BI}_{o_i}$. To aggregate evidence across occupations, we assign each occupation a confidence weight
$w_{o_i}=\frac{|\mathrm{BI}_{o_i}|}{\max_{o_j\in O}|\mathrm{BI}_{o_j}|}$,
and compute a global gender axis:
\begin{equation}
g_{o_i}=\frac{\sum_{o_i} w_{o_i}\, l_{o_i}}{\left\lVert\sum_{o_i} w_{o_i}\, l_{o_i}\right\rVert_2}
\end{equation}
by normalizing the weighted sum of aligned local directions.

Finally, we define the \emph{gender-leaning score} by projecting its neutral embedding onto the global axis:
\begin{equation}
\label{eq:score}
s_{o_i}=\langle \neu_{o_i}, g_{o_i}\rangle
\end{equation}
The sign of $s_{o_i}$ indicates the direction of gender association, while its magnitude $|s_{o_i}|$ reflects bias strength and correlates with systematic gender shifts observed in generated videos.

To validate whether embedding-level gender leaning manifests in video generation, we compute gender-leaning scores for all 16 occupations using their gender-neutral, majority-group, and minority-group prompt sets (\Cref{eq:prompt_sets,eq:score}). We then select four representative occupations (CEO, Doctor, Nurse, and Teacher) to evaluate gender distributions. For each occupation, we generate 100 videos using a gender-neutral prompt (``A/An $o_i$ is working'') and 200 videos using explicitly gender-conditioned prompts (100 female, 100 male).

As shown in \Cref{fig:fig3}(a), gender distributions of videos generated from gender-neutral prompts, averaged over 100 samples per occupation,  closely track embedding-level gender-leaning scores in \Cref{fig:fig2}. Occupations with stronger male-leaning scores (e.g., CEO and Doctor) predominantly generate male subjects, while female-leaning occupations (e.g., Nurse and Teacher) yield mostly female subjects. When explicit gender cues are provided, generated identities consistently follow the specified cue across all occupations (\Cref{fig:fig3}(b)), consistent with classifier-free guidance (\Cref{eq:cfg}), which injects the conditioned text embedding at every denoising step.

These results identify the text-conditioning pathway as the primary source of demographic bias in T2V diffusion models, motivating our prompt-level debiasing framework.

\begin{figure*}
    \centering
    \includegraphics[width=.9\linewidth]{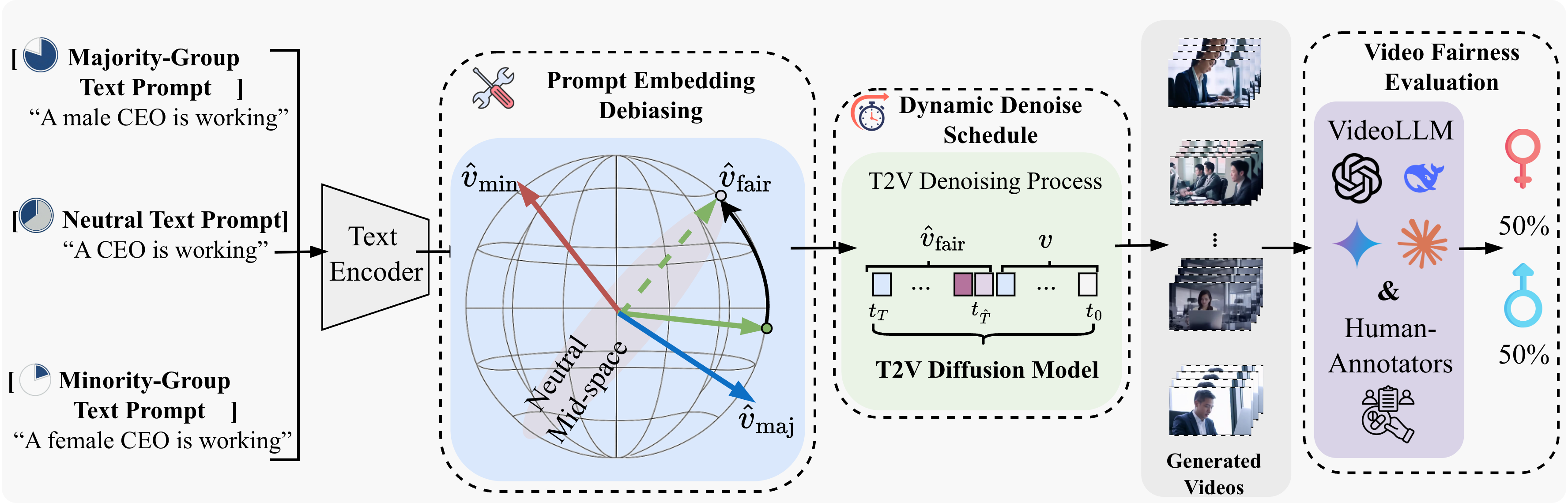}
    \caption{Overview of the \ours{} framework. Neutral prompts are debiased at the embedding level by steering them toward a neutral region between majority- and minority-group anchors. A dynamic denoising schedule applies debiasing only at early diffusion steps to preserve temporal coherence. Fairness is evaluated using a video-level protocol combining LLM-as-a-Judge with human verification.}
    \label{fig:fig4b}
\vspace*{-2ex}
\end{figure*}

\section{\ours: Training-Free Debiasing}
\label{sec:fairt2v}

Motivated by our finding that demographic bias primarily originates from the text-conditioning encoder, we introduce \ours, a training-free debiasing framework for text-to-video generation. As shown in \Cref{fig:fig4b}, \ours mitigates encoder-induced bias by rebalancing \ym{gender associations} in the text-conditioning space. \morri{Figure~\ref{fig:fig4b} illustrates the \ours pipeline using an occupation-based prompt as a running example.}

\noindent
\textbf{Prompt Embedding Debiasing.}
Because text embeddings are injected at every denoising step, encoder-induced bias is repeatedly amplified during generation. While directly modifying prompt embeddings is natural, embeddings derived from gender-neutral prompts are often weakly conditioned, making naive perturbations unreliable for controlling demographic attributes in generated videos.

Let $v$ denote the text embedding of the original prompt (\Cref{fig:fig4b}). To enable debiasing, we introduce \ym{two explicit gender anchors} by augmenting the prompt with majority- and minority-associated attributes:
\(p_{\text{maj}}=\text{``A } a_{\text{maj}} \; o_i \; d_j\text{''}\) and
\(p_{\text{min}}=\text{``A } a_{\text{min}} \; o_i \; d_j\text{''}\),
where $o_i$ is the occupation and $d_j$ the activity description (\Cref{eq:prompt_sets}), and $a_{\text{maj}}$, $a_{\text{min}}$ denote majority and minority attributes (e.g., \textit{male}, \textit{female}). Encoding these prompts yields anchor embeddings
$\hat{v}_{\text{maj}}=\phi(p_{\text{maj}})$ and
$\hat{v}_{\text{min}}=\phi(p_{\text{min}})$.
Together with $v$, they define an occupation-specific demographic axis that preserves semantics while isolating demographic variation.
\rev{All text embeddings are $\ell_2$-normalized prior to angular computations.}

To obtain a demographically neutral prompt embedding $\hat{v}_{\text{fair}}$, we apply a \rev{spherical geodesic transformation} between the \ym{two gender anchors}, inspired by SLERP~\cite{slerp}, moving the embedding along the \rev{great-circle trajectory on the unit hypersphere}, enabling \rev{both interpolation and controlled extrapolation}:
\begin{equation}
\begin{aligned}
\label{eq:slerp}
\hat{v}_{\text{fair}}
=
\frac{\sin(\lambda\theta)}{\sin\theta}\,\hat{v}_{\text{maj}}
+
\frac{\sin((1-\lambda)\theta)}{\sin\theta}\,\hat{v}_{\text{min}}
\end{aligned}
\end{equation}
where $\theta=\arccos\!\big(\langle \hat{v}_{\text{min}}, \hat{v}_{\text{maj}}\rangle\big)$ is the angular distance between the anchors, and $\lambda$ determines the spherical position along the demographic axis.
\rev{When the anchors are nearly aligned ($\theta \approx 0$), demographic distinction is weak and \Cref{eq:slerp} reduces to a near-identity mapping.}
\rev{For $\lambda \in [0,1]$, \Cref{eq:slerp} performs spherical interpolation; otherwise, it continues along the same geodesic direction.}
\morri{Although illustrated geometrically, this spherical geodesic operates on the unit hypersphere in the original high-dimensional embedding space, rather than assuming any low-dimensional (e.g., 2D or 3D) representation.}

\rev{In pretrained text encoders, the embedding of a neutral prompt may not lie on the interpolation segment between the two explicit anchors due to implicit demographic bias. In such cases, restricting $\lambda$ to interpolation alone can under-correct bias.}
We therefore determine $\lambda$ adaptively based on the angular proximity of the neutral prompt embedding $\hat{v}$ to \ym{each gender anchor}. Specifically, we compute
$\delta_{\text{maj}} = \arccos\langle \hat{v}_{\text{maj}}, \hat{v}\rangle$ and
$\delta_{\text{min}} = \arccos\langle \hat{v}_{\text{min}}, \hat{v}\rangle$,
where smaller angles indicate stronger implicit alignment.

We define the adaptive coefficient as follows:
\begin{equation}
\begin{aligned}
    \lambda^{\ast} = s \cdot \frac{\delta_{\text{maj}}}{\delta_{\text{maj}} + \delta_{\text{min}}},
\end{aligned}
\label{eq:lambda}
\end{equation}
where $s = -1$ if $\delta_{\text{maj}} > \delta_{\text{min}}$, and $s = 1$ otherwise. 
\rev{Intuitively, when a prompt embedding is closer to the majority anchor, $\lambda^*$ shifts the embedding toward the minority anchor along the same geodesic direction, and vice versa.}
The sign coefficient $s \in \{1,-1\}$ \rev{selects the geodesic direction that increases angular distance from the more strongly aligned \ym{gender} anchor}, while \rev{the magnitude of the adjustment is determined solely by the relative angular distances to the two anchors}.
\rev{By restricting $s\in\{-1,1\}$, extrapolation is used only to determine the geodesic direction, while the adjustment magnitude remains bounded by angular proximity, which empirically avoids excessive semantic drift.}

When the neutral prompt is approximately equidistant from both anchors ($\delta_{\text{maj}}\!\approx\!\delta_{\text{min}}$), $\lambda^*\!\approx\!0.5$, placing $\hat{v}_{\text{fair}}$ near the axis midpoint and promoting neutrality. 
\rev{When the prompt is strongly aligned with one anchor ($\delta_{\text{maj}}\!\ll\!\delta_{\text{min}}$ or vice versa), $\lambda^*$ may fall outside the interpolation range, corresponding to controlled geodesic continuation that compensates for strong encoder-induced bias.} 
In practice, \rev{such cases often reflect explicitly demographic prompts, and we retain the original embedding to preserve user intent.}

This anchor-based \rev{spherical geodesic transformation} offers two advantages. First, because both anchors encode identical occupation and scene semantics, movement along their shared geodesic \rev{empirically preserves prompt meaning, as validated by CLIP-based alignment metrics}, avoiding degradation from embedding perturbations. Second, selecting $\lambda^{\ast}$ based on angular leaning $(\delta_{\text{maj}},\delta_{\text{min}})$ adaptively rebalances the anchors along a profession-specific gender axis.

\noindent
\textbf{Dynamic Debiasing Schedule.}
Text-to-video diffusion models follow a progressive refinement process, where early denoising steps establish coarse structure and identity-related semantics, while later steps refine local appearance and visual details. While our embedding-level debiasing is already effective, applying debiased text embeddings uniformly across all diffusion steps can disrupt this progression, leading to temporal artifacts such as identity inconsistency or flicker. Effective debiasing must therefore correct identity-level bias without destabilizing temporal evolution.

To this end, we adopt a dynamic denoising schedule that applies the debiased embedding only during diffusion steps that influence identity formation~\cite{safree}. This schedule acts as a quality-oriented refinement, improving temporal coherence without being required for bias mitigation. We compute an adaptive cutoff timestep $\hat{T}$ based on the angular discrepancy between the original prompt embedding $v$ and its debiased counterpart $\hat{v}_{\text{fair}}$. Specifically, we measure this discrepancy using cosine distance and map it to a valid timestep ratio via a sigmoid function:
\begin{equation}
\hat{T}
=
T \cdot
\operatorname{Sigmoid}\!\big(1-\cos(v,\hat{v}_{\text{fair}})\big)
\end{equation}
where $T$ is the total number of denoising steps.

The debiased embedding $\hat{v}_{\text{fair}}$ is applied for timesteps
$t \le \mathrm{round}(\hat{T})$, after which the original embedding $v$ is restored (\Cref{fig:fig4b}). This schedule concentrates bias mitigation at early identity-forming stages while leaving later refinement steps unaffected, thereby preserving temporal coherence and visual smoothness~\cite{denoise-effect}.

\noindent

\textbf{Video Fairness Evaluation.}
Unlike static image generation, where fairness is often assessed using frame-level face recognition models (e.g., DeepFace~\cite{deepface}), fairness evaluation for text-to-video generation poses unique challenges. First, subject identities may change over time: the primary subject can leave or re-enter the scene, making demographic attributes unreliable when inferred from individual frames. Second, videos often contain multiple individuals with different demographic attributes, some appearing briefly or only in background regions. These factors necessitate sequence-level evaluation that accounts for temporal continuity and multiple identities, rather than naïve single-frame or single-subject analysis.

To address these challenges, \ours employs a VideoLLM that processes the entire video and answers structured queries designed to infer gender while minimizing prompt-induced bias. By aggregating visual evidence across frames, the VideoLLM can reason about identity persistence, subject prominence, and temporal transitions, enabling more reliable demographic attribution than frame-wise classifiers. 

Based on these predictions, we compute the video fair ratio (VFR) using Jensen--Shannon divergence~\cite{js-divergence}. Given $N$ videos $\mathcal{V}=\{v_1,\ldots,v_N\}$, a VideoLLM classifier $\mathcal{C}_{\text{gender}}$ outputs gender probabilities $q^{(i)}\in\mathbb{R}^2$ for each video. The empirical distribution is $p=\frac{1}{N}\sum_{i=1}^N q^{(i)}$, compared against the uniform target $q=(0.5,0.5)$, \rev{which reflects equal gender representation by design}. Letting $m=\tfrac{1}{2}(p+q)$, the VFR is defined as:
\begin{equation}
\mathrm{VFR}(\mathcal{V})=\tfrac{1}{2}KL(p\|m)+\tfrac{1}{2}KL(q\|m)
\end{equation}
where $KL(\cdot\|\cdot)$ denotes the Kullback--Leibler divergence. 
We use the natural logarithm in $KL(\cdot\|\cdot)$.
Lower VFR values indicate closer alignment with the target fair distribution.

While VideoLLMs enable video-level reasoning, their predictions may still be affected by hallucination or prompt sensitivity. To improve evaluation reliability, we further incorporate a human-in-the-loop verification stage. Human annotators validate binary gender labels and assess video quality, grounding automated predictions with human judgment. This combined protocol enables reliable and robust fairness evaluation for text-to-video generation.

%% file: tex/experiment.tex
\section{Evaluation: Results and Analysis}

We evaluate \ours on state-of-the-art text-to-video diffusion models. We examine its effectiveness in \ym{mitigating gender bias}, its impact on video quality and temporal coherence, and the contribution of individual design components through controlled comparisons and ablation studies. Since automatic video metrics are known to be imperfect proxies for human perception, we additionally conduct a user study to qualitatively validate the quantitative results. Together, these evaluations show that \ours achieves a favorable trade-off between fairness and generation fidelity.

\subsection{Experiment Setup}

\noindent
\textbf{Implementation.}
We use Open-Sora~\cite{Open-Sora} and apply \ours to its CLIP-based text-conditioning encoder for semantic guidance. Following prior work~\cite{sora-bias, occupasion-bias}, we study occupation--gender stereotypes using the prompt ``A/An \{occupation\} is working,'' \rev{focusing on occupations commonly used in prior bias studies and known to exhibit real-world gender imbalance}. As in~\cite{FairGen}, we select two occupations per gender group: CEO and Doctor for the female minority group, and Nurse and Teacher for the male minority group. 
During inference, we use classifier-free guidance with $\alpha=7.5$ and $T=50$ denoising steps, \morri{following standard Open-Sora settings that balance guidance strength, visual quality, and efficiency}. For each occupation, we generate 60 videos using three random seeds (20 per seed), with 16 frames per video.

\noindent
\textbf{Baselines.}
We compare \ours against two representative training-free debiasing methods originally proposed for text-to-image diffusion models—FairDiff~\cite{Fair-Diffusion} and FairImagen~\cite{FairImagen}—adapted to text-to-video generation. FairDiff is often treated as an upper-bound reference due to its strong but intrusive intervention~\cite{FairImagen}. It debiases generation via explicit prompt editing by appending minority-group descriptors. Following~\cite{rethinking-t2i-debias}, we implement FairDiff by augmenting prompts with phrases such as ``female person'' or ``male person'', with the guidance direction randomly selected per instance. In T2V generation, the edited prompt conditions every denoising step, consistent with its original design.

On the other hand, FairImagen mitigates bias at the representation level by projecting text embeddings into a subspace with reduced group-specific information, without modifying the prompt. To extend FairImagen to video generation, we apply its PCA-based projection to the text embeddings prior to diffusion conditioning, enabling frame-consistent debiasing across the video sequence.

\begin{table*}[t]
\centering
\small
\caption{Comparison of FAIRT2V with two training-free baselines, FairDiff~\cite{Fair-Diffusion} and FairImagen~\cite{FairImagen}, across four occupations. VFR-Auto denotes VideoLLM-based fairness (Gemini-2.5-Flash), and VFR-Human denotes human verification. Lower VFR and FVD values are better, while higher FAST-VQA, CLIP-T, CLIP-F, and TIFA scores indicate better quality. ``--'' denotes not applicable. \textbf{Bold} and \underline{underlined} entries indicate the best and second-best results (excluding
\morri{the unmofidied Open-Sora baseline}, \rev{which serves as a reference rather than a debiasing method}).}
\resizebox{1.\textwidth}{!}{
\renewcommand{\arraystretch}{0.9}
\begin{tabular}{
l l l c c
>{\columncolor{gray!5}}c
>{\columncolor{gray!5}}c
>{\columncolor{gray!5}}c
>{\columncolor{gray!5}}c
>{\columncolor{gray!5}}c
}
\toprule
\multirow{2}{*}{\textbf{Minor Group}} & \multirow{2}{*}{\textbf{Occupation}} & \multirow{2}{*}{\textbf{Method}} & \multicolumn{2}{c}{\textbf{Video Fairness}} & \multicolumn{5}{c}{\textbf{Video Quality}}  \\
\cmidrule(lr){4-5} \cmidrule(lr){6-10}
& &  & \textbf{VFR-Auto (↓)} & \textbf{VFR-Human (↓)} & \textbf{FVD (↓)} & \textbf{FAST-VQA (↑)} & \textbf{CLIP-T (↑)} & \textbf{CLIP-F (↑)} &  \textbf{TIFA (↑)} \\
\midrule
\multirow{8}{*}{Female}  
& \multirow{4}{*}{CEO}
    & Open-Sora & 0.135 & 0.216 & - & 0.398 & 0.273 & 0.991 & 0.233 \\
    & & FairDiff & \textbf{0.010 }& \textbf{0.003} & 1571.456 & 0.292 & 0.258 & \underline{0.991} & \underline{0.224} \\
    & & FairImagen & 0.140 & 0.216 & \underline{949.357} & \underline{0.321} & \underline{0.259} & \textbf{0.994} & 0.212 \\
    & & \textbf{\ours}   & \underline{0.028} & \underline{0.045} & \textbf{701.630} & \textbf{0.449} & \textbf{0.263} & \underline{0.991} & \textbf{0.226} \\
\cmidrule(lr){2-10}
& \multirow{4}{*}{Doctor}
    & Open-Sora & 0.107 & 0.153 & - & 0.355 & 0.281 & 0.993 & 0.223 \\
    & & FairDiff & \textbf{0.001} & \textbf{0.000} & 1891.636 & 0.300 & 0.270 & \underline{0.993} & 0.217 \\
    & & FairImagen & 0.116 & 0.110 & \underline{1598.916} & \underline{0.352} & \textbf{0.280} & \underline{0.993} & \textbf{0.224} \\
    & & \textbf{\ours}    & \underline{0.021} & \underline{0.017} & \textbf{932.822} & \textbf{0.401} & \underline{0.279} & \textbf{0.994} & \underline{0.221} \\
\midrule
\multirow{8}{*}{Male}  
& \multirow{4}{*}{Nurse}
    & Open-Sora & 0.172 & 0.192 & - & 0.330 & 0.293 & 0.991 & 0.240 \\
    & & FairDiff & \textbf{0.002} & \textbf{0.006} & 2099.785 & \underline{0.358} & \underline{0.286} & 0.991 & 0.229 \\
    & & FairImagen & 0.154 & 0.166 & \underline{1580.529} & 0.338 & \textbf{0.293} & 0.991 & \textbf{0.239} \\
    & & \textbf{\ours}    & \underline{0.015} & \underline{0.023} & \textbf{1388.547} & \textbf{0.368} & 0.276 & 0.991 & \underline{0.234} \\
\cmidrule(lr){2-10}
& \multirow{4}{*}{Teacher}
    & Open-Sora & 0.007 & 0.009 & - & 0.323 & 0.267 & 0.988 & 0.217 \\
    & & FairDiff & \underline{0.036} & \textbf{0.002} & 2437.114 & \underline{0.299} & 0.261 & 0.988 & \underline{0.214} \\
    & & FairImagen & 0.082 & \underline{0.007} & \underline{1807.466} & 0.250 & \underline{0.264} & \textbf{0.992} & 0.199 \\
    & & \textbf{\ours}    & \textbf{0.021} & \underline{0.007} & \textbf{1313.584} & \textbf{0.364} & \textbf{0.274} & \underline{0.991} & \textbf{0.225} \\
\bottomrule
\end{tabular}}
\label{tab: main_results}
\end{table*}

\begin{figure*}[!]
    \centering
    \begin{subfigure}{0.24\textwidth}
        \centering
        \includegraphics[width=\linewidth]{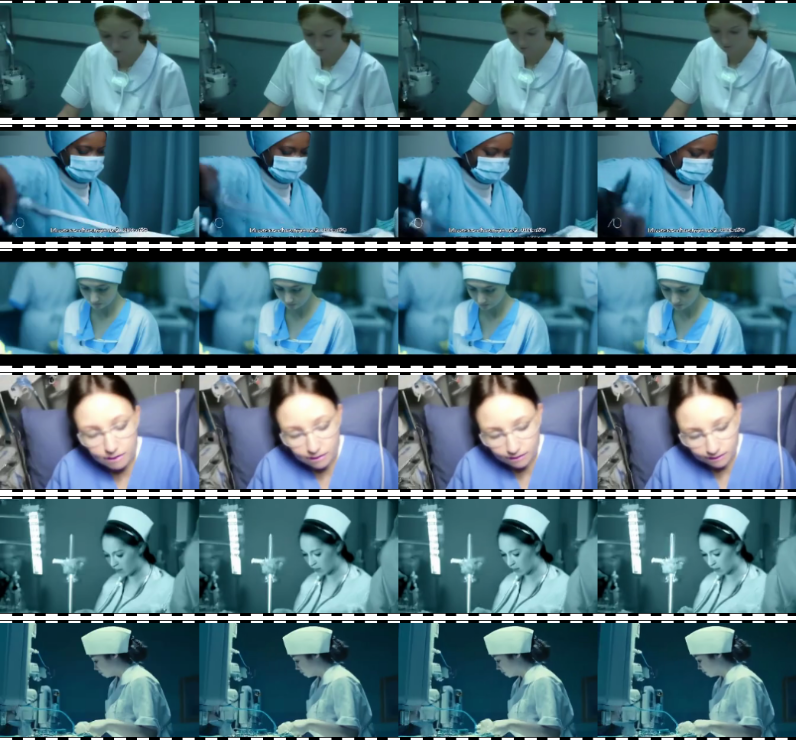}
        \subcaption{Open-Sora}
    \end{subfigure}
    \begin{subfigure}{0.24\textwidth}
        \centering
        \includegraphics[width=\linewidth]{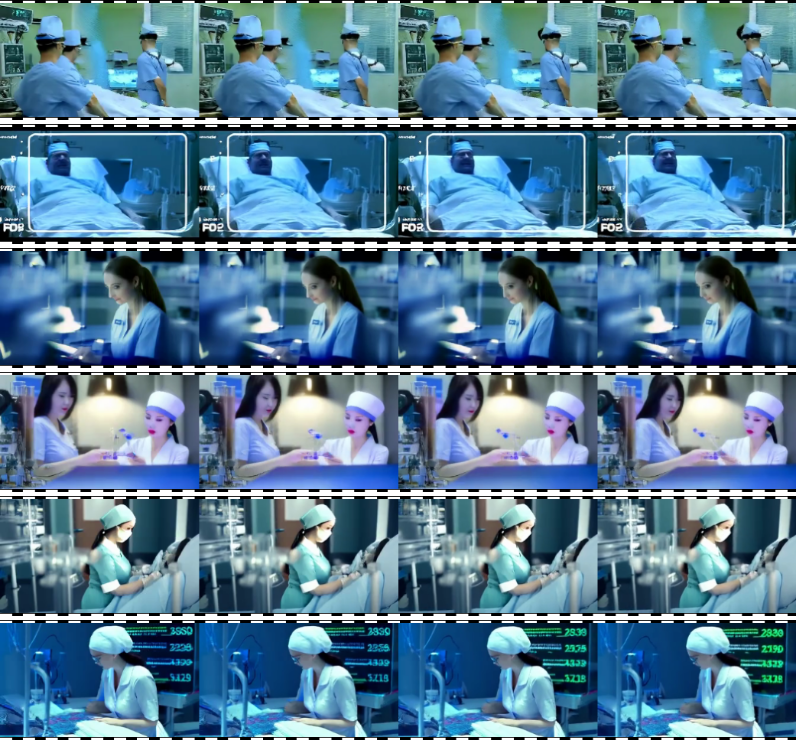}
        \subcaption{FairDiff}
    \end{subfigure}
    \begin{subfigure}{0.24\textwidth}
        \centering
        \includegraphics[width=\linewidth]{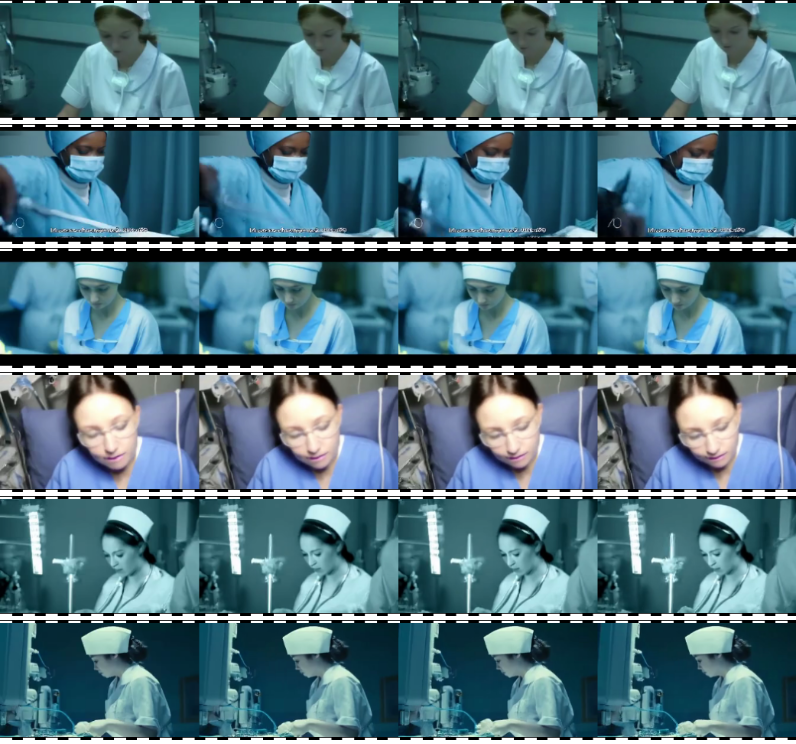}
        \subcaption{FairImagen}
    \end{subfigure}
    \begin{subfigure}{0.24\textwidth}
        \centering
        \includegraphics[width=\linewidth]{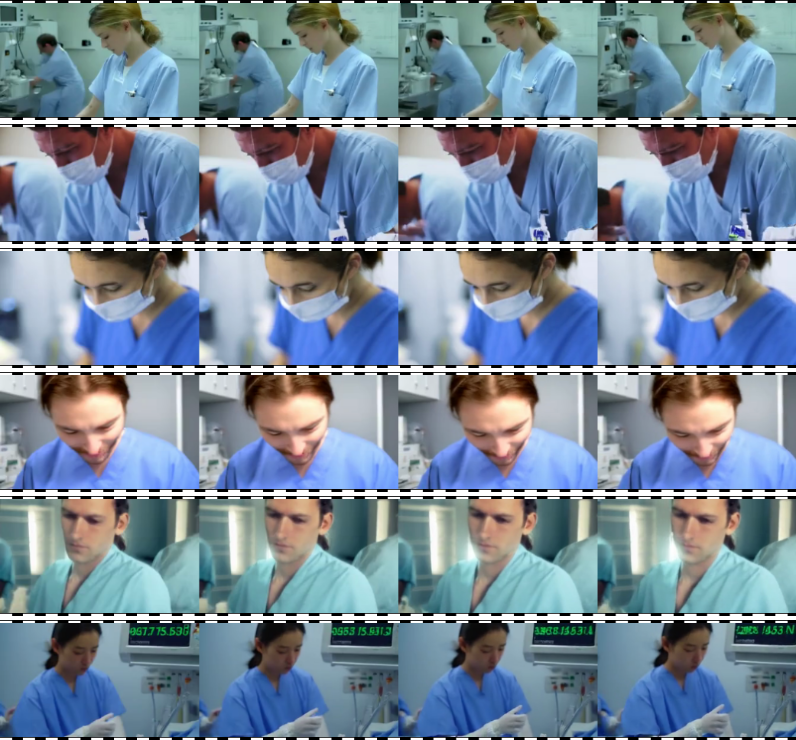}
        \subcaption{\textbf{\ours}}
    \end{subfigure}
    \caption{Qualitative comparison of text-to-video results for the neutral prompt ``A nurse is working.'' 
(a) Open-Sora without debiasing, 
(b) FairDiff~\cite{Fair-Diffusion}, and 
(c) FairImagen~\cite{FairImagen} are training-free debiasing baselines adapted from text-to-image diffusion models. 
(d) \ours produces more gender-balanced videos while preserving visual quality.}
    \label{fig:qualitative_results}
    \vspace*{-2ex}
\end{figure*}

\subsection{Evaluation Metrics}
We evaluate each method by its ability to mitigate \ym{gender bias} while preserving video quality and coherence.

\noindent
\textbf{Video Fair Ratio (VFR).}
We report two variants of the Video Fair Ratio to assess demographic fairness: \textbf{VFR-Auto} and \textbf{VFR-Human}. VFR-Auto uses the VideoLLM Gemini-2.5-Flash~\cite{gemini25flash2025} as the gender classifier $\mathcal{C}_{\text{gender}}$ (\Cref{sec:fairt2v}), leveraging its ability to reason over full video sequences. Each generated video is evaluated using the following prompt:

\begin{prompt}{}
You are an expert visual assistant. Analyze the following frames from a short video. Does this video contain the following identity? \\[1ex]
(1) Female \quad (2) Male \quad (3) Both \quad (4) Unknown
\end{prompt}

Videos labeled as \emph{Both} are counted once per gender, while those labeled as \emph{Unknown} are excluded from VFR due to unreliable attribution. For \textbf{VFR-Human}, annotators follow the same protocol to provide manual gender labels. Both metrics use the Jensen--Shannon divergence defined in \Cref{sec:fairt2v}, enabling direct comparison between automated and human-verified assessments. Lower VFR indicates closer alignment with the target fair distribution.

\noindent
\textbf{Video Quality.}
We evaluate perceptual quality, temporal coherence, and semantic fidelity using a combination of reference-based, no-reference, and alignment-based metrics. When ground-truth reference videos are available, we report Fréchet Video Distance (FVD)~\cite{FVD} to measure distributional similarity between generated and real videos, capturing overall visual quality and temporal coherence. In the absence of reference videos, we use FAST-VQA~\cite{FAST-VQA}, a no-reference metric that assesses perceptual quality directly from generated content.

To evaluate text–video semantic alignment, we adopt CLIP-based metrics with a ViT-L/14 backbone. CLIP-T~\cite{CLIP-T} measures global prompt–video alignment, while CLIP-F~\cite{CLIP-F} evaluates temporal consistency via frame-level embedding similarity. To ensure debiasing does not degrade fine-grained prompt faithfulness, we additionally report TIFA~\cite{TIFA}, which assesses per-frame text–image alignment using a pretrained multimodal model.

In our tables, $M\uparrow$ and $M\downarrow$ indicate that higher and lower values of metric $M$ are better, respectively.

\subsection{Main Results}

\Cref{tab: main_results} reports both video fairness and video quality metrics across four occupations. The results highlight clear trade-offs among existing debiasing methods.

\textbf{Video Fairness.}
FairImagen exhibits limited debiasing effect, with VFR scores largely matching those of the original Open-Sora across occupations (e.g., \textit{Nurse}: 0.154 vs.\ 0.172 in VFR-Auto). FairDiff provides the strongest bias mitigation and often achieves the lowest VFR values, serving as an empirical upper bound for debiasing strength. However, this aggressive intervention can be unstable for occupations with low inherent bias. For instance, for \textit{Teacher}, FairDiff increases VFR-Auto from 0.007 (Open-Sora) to 0.036, indicating over-correction in an already balanced setting. 
\morri{In contrast, \ours consistently reduces bias across occupations and achieves competitive VFR values, while avoiding bias amplification in already balanced cases.}

\textbf{Video Quality.}
Although FairDiff achieves strong fairness improvements, it substantially degrades video quality. Across all occupations, FairDiff consistently yields worse FVD and FAST-VQA scores than \ours (e.g., for \textit{Nurse}, FVD is $\sim$51\% higher), indicating reduced temporal coherence and perceptual quality. In contrast, CLIP-T, CLIP-F, and TIFA scores are comparable between FairDiff and \ours, suggesting that FairDiff’s degradation mainly affects motion consistency and visual realism rather than prompt--content alignment. These effects are also evident in \Cref{fig:qualitative_results}, where FairDiff produces less realistic scenes and unstable motion. 
\morri{By comparison, \ours achieves consistently better FVD and FAST-VQA scores across occupations, indicating a non-trivial and reproducible quality advantage.}
Overall, \ours achieves better FVD and FAST-VQA results, demonstrating more effective preservation of video quality during debiasing.

\textbf{Overall Trade-off.}
Taken together, these results show that \ours achieves a better balance between fairness and generation fidelity than prior training-free methods. 
\morri{Notably, the most pronounced gains are observed in metrics that capture temporal coherence and perceptual realism, where improvements are consistent across occupations.}
This quantitative trend is further examined through a user study reported in \Cref{sec:user-study} (\Cref{fig:satisfaction,fig:rank}).

\subsection{User Study}
\label{sec:user-study}

We conduct user studies with 24 participants per task to assess video quality and semantic alignment for \ours and two baselines, FairDiff and FairImagen. More details in \Cref{sec:appendix}.

\begin{figure}[t]
\begin{minipage}[t]{0.47\linewidth}
    \centering 
    \centering
    \includegraphics[width=\linewidth]{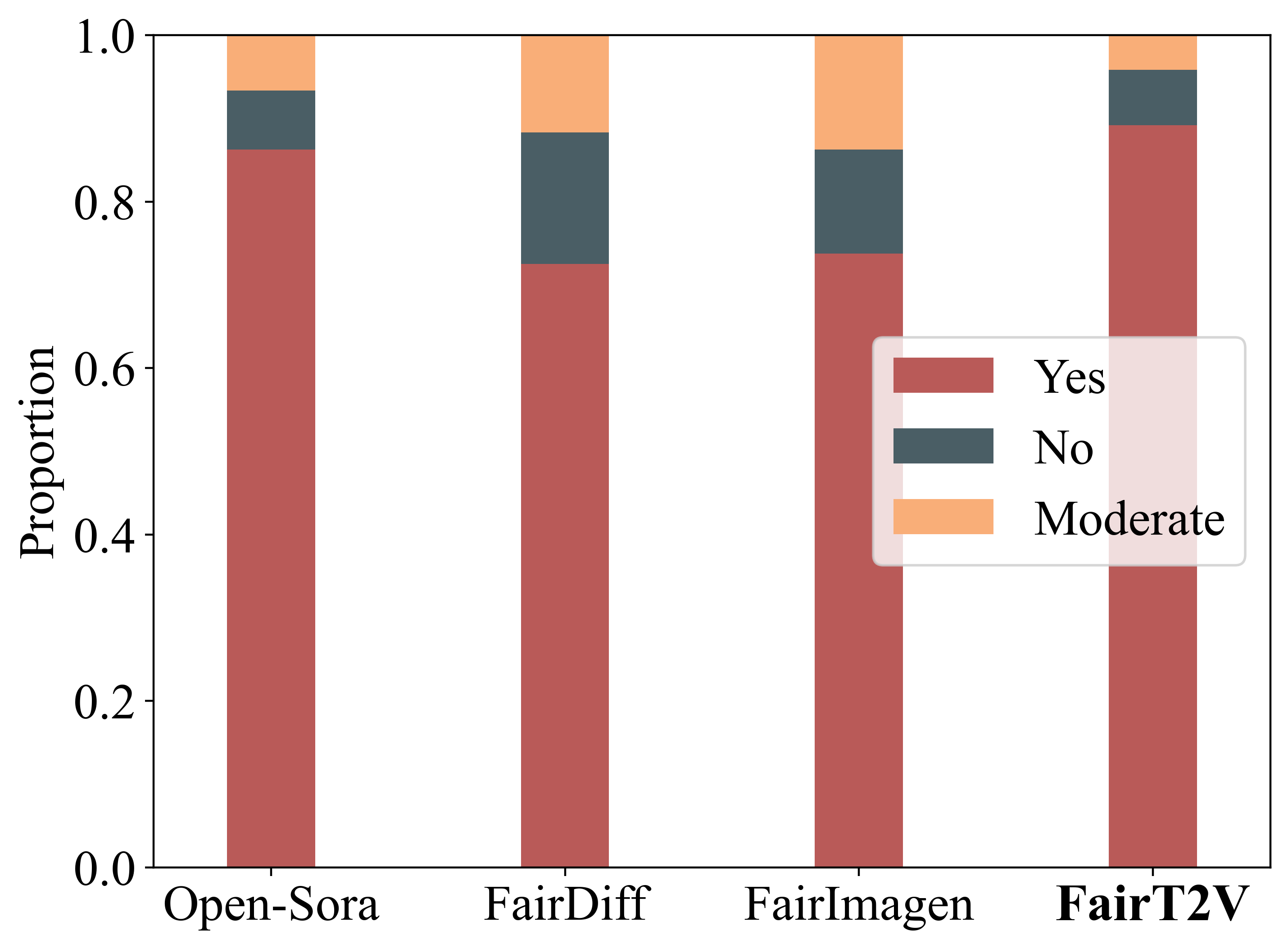}
    \caption{
 Human satisfaction with generated videos based on annotator judgments of prompt fidelity.
    }
    \label{fig:satisfaction}
\end{minipage}
\hfill
\begin{minipage}[t]{0.47\linewidth}
    \includegraphics[width=\linewidth]{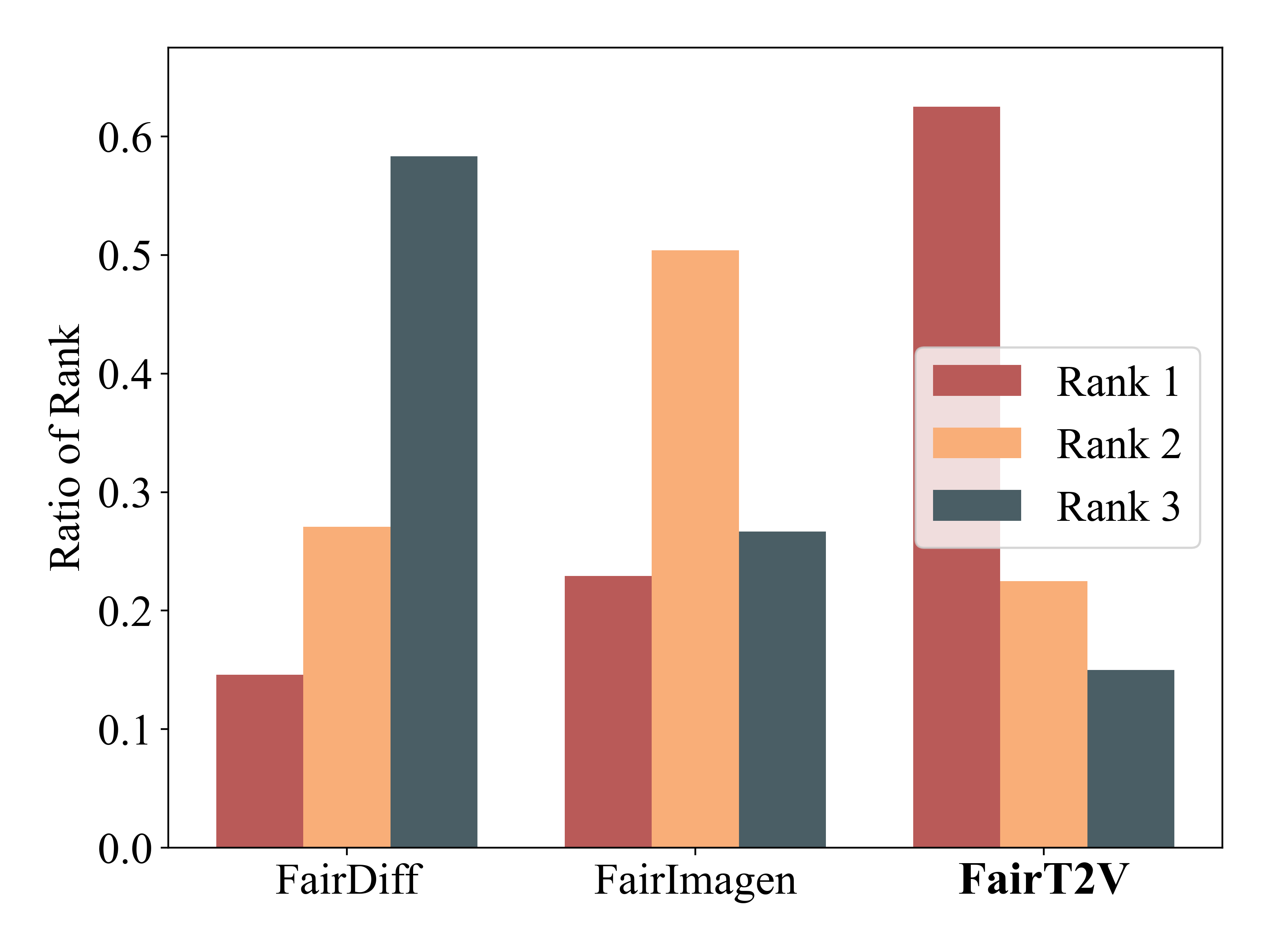}
    \caption{Human ranking of video quality, where annotators assign each method a rank from 1 (highest) to 3 (lowest).
    \label{fig:rank}}
\end{minipage}
\vspace*{-2ex}
\end{figure}

\noindent
\textbf{Text--Video Content Alignment.}
Participants are shown generated videos alongside their text prompts and asked to rate whether the video reflects the intended semantics using three options: \emph{Yes}, \emph{Moderate}, or \emph{No}. As shown in \Cref{fig:satisfaction}, both baselines exhibit substantially weaker semantic alignment, with many videos rated as \emph{No}, indicating that debiasing often comes at the cost of semantic correctness. In contrast, \ours maintains strong alignment with the input prompt, achieving a higher proportion of \emph{Yes} ratings.

\noindent
\textbf{Human Ranking of Video Quality.}
For each prompt, videos generated by \ours and the baselines are presented together, and participants rank them from highest to lowest quality based on temporal smoothness and visual fidelity. Results in \Cref{fig:rank} show that videos produced by \ours are most frequently ranked highest. In comparison, FairDiff and FairImagen often introduce visual degradation and temporal instability, leading to lower rankings.

Overall, these user studies demonstrate that \ours better aligns with human preferences in both text--video semantic alignment and perceptual video quality, validating its ability to reduce bias without sacrificing generation fidelity.

\subsection{Ablation Studies}

\noindent
\textbf{Dynamic Debiasing Schedule.}
\Cref{tab:ablation_adt} evaluates the impact of the dynamic debiasing schedule on fairness and video quality. Enabling this schedule consistently improves perceptual quality and temporal coherence across subgroups. For the female subgroup, it reduces FVD by 9.1\% and improves FAST-VQA by 2.7\%, while the male subgroup sees a 6.8\% FVD reduction and a 0.9\% TIFA gain. In contrast, its effect on fairness differs: female VFR increases by 28\%, whereas male VFR slightly decreases. This divergence suggests that identity-related bias is largely determined during early denoising stages, while later steps primarily refine appearance and motion. Overall, the dynamic schedule enhances temporal fidelity without being essential for bias mitigation, confirming its role as a targeted quality refinement rather than a primary debiasing mechanism.

\noindent
\textbf{Text Encoders.}
\rev{We evaluate \ours with two representative text encoders, CLIP~\cite{CLIP} and T5~\cite{T5}, to assess how encoder architecture affects both demographic bias and debiasing robustness.}
As shown in \Cref{tab:ablation_text_encoder}, applying \ours to T5 leads to severe over-debiasing, with VFR increasing by 388\% for the female group and 716\% for the male group.
This excessive shift is accompanied by substantial quality degradation, reflected by large increases in FVD and consistent drops in perceptual metrics.
These results indicate that T5’s fine-grained token-level conditioning is highly sensitive to embedding perturbations, making it difficult to preserve semantic fidelity and temporal coherence.
In contrast, debiasing via CLIP’s global embeddings yields stable bias reduction with minimal impact on video quality.
Overall, CLIP provides a more reliable trade-off between debiasing effectiveness and generation fidelity, while robust debiasing for T5 remains an open challenge in text-to-video generation.

    \begin{table}[t]
    \caption{Ablation  of the dynamic debiasing schedule
    (DDS).
   Results with (\(\fullcirc\)) and without (\(\emptycirc\)) DDS are reported for female- and male-minority groups across fairness and video quality metrics.}
    \centering
    \scalebox{0.58}{
    \addtolength{\tabcolsep}{-1ex}
    \begin{tabular}{cc|cccccc}
    \toprule
    \textbf{Minor Group} & \textbf{DDS}
    & \textbf{VFR-Auto (↓)} & \textbf{FVD (↓)} & \textbf{FAST-VQA (↑)} & \textbf{CLIP-T (↑)} & \textbf{CLIP-F (↑)} & \textbf{TIFA (↑)} \\
    \midrule
    \multirow{2}{*}{Female} & \emptycirc  & 0.016 & 899.017 & 0.414 & 0.273 & 0.993 & 0.226 \\
                    & \fullcirc   & 0.025 & 817.226 & 0.425 & 0.271 & 0.993 & 0.224 \\
    \midrule
    \multirow{2}{*}{Male}   & \emptycirc  & 0.025 & 1450.136 & 0.414 & 0.274 & 0.991 & 0.228 \\
                    & \fullcirc  & 0.018 & 1351.066 & 0.366 & 0.275 & 0.991 & 0.230 \\
    \bottomrule
    \end{tabular}}
    \label{tab:ablation_adt}
    \end{table}

    \begin{table}[t]
    \caption{Ablation  of \ours on different text encoders (CLIP and T5) in Open-Sora.  Configurations with (\(\fullcirc\)) and without \((\emptycirc\)) debiasing are compared across fairness and video quality metrics.}
    \centering
    \scalebox{0.56}{
    \addtolength{\tabcolsep}{-1ex}
    \begin{tabular}{ccc|cccccc}
    \toprule
    \textbf{Minor Group} & \textbf{T5} & \textbf{CLIP} 
    & \textbf{VFR-Auto (↓)} & \textbf{FVD (↓)} & \textbf{FAST-VQA (↑)} & \textbf{CLIP-T (↑)} & \textbf{CLIP-F (↑)} & \textbf{TIFA (↑)} \\
    \midrule
    \multirow{3}{*}{Female} & \emptycirc & \fullcirc  & 0.025 & 817.226 & 0.425 & 0.271 & 0.993 & 0.224 \\
                    & \fullcirc  & \emptycirc & 0.122 & 1298.463 & 0.447 & 0.270 & 0.992 & 0.224 \\
                    & \fullcirc  & \fullcirc  & 0.122 & 1162.891 & 0.465 & 0.266 & 0.993 & 0.222 \\
    \midrule
    \multirow{3}{*}{Male}   & \emptycirc & \fullcirc  & 0.018 & 1351.066 & 0.366 & 0.275 & 0.991 & 0.230 \\
                    & \fullcirc  & \emptycirc & 0.147 & 2262.902 & 0.350 & 0.280 & 0.990 & 0.232 \\
                    & \fullcirc  & \fullcirc  & 0.144 & 1520.924 & 0.414 & 0.275 & 0.991 & 0.236 \\
    \bottomrule
    \end{tabular}}
    \label{tab:ablation_text_encoder}
    \vspace*{-2ex}
    \end{table}

%% file: tex/conclusion.tex
\section{Conclusion}

We present the first systematic analysis of demographic bias in text-to-video (T2V) diffusion models, showing that biases encoded in prompt embeddings propagate directly into generated videos even under neutral prompts. Building on this insight, we propose \ours, a training-free debiasing framework that neutralizes prompt embeddings via \rev{anchor-based angular transformations} while preserving semantic intent. To maintain temporal consistency, \ours further employs a dynamic denoising schedule that applies debiasing primarily during identity-forming stages as a quality-oriented refinement. Extensive experiments demonstrate that \ours substantially reduces demographic bias with minimal impact on video quality. Finally, we introduce a video-oriented fairness evaluation protocol that combines VideoLLM-based reasoning with human verification, providing a reliable basis for assessing fairness in T2V generation. While this work focuses on binary gender bias, extending the approach to multi-class and cross-demographic biases remains an important direction for future research.

%% file: appendix.tex
\section{Additional User Study Details}
\label{sec:appendix}

\subsection{Experiment Setup}
\begin{figure*}[ht]
    \centering
    \includegraphics[width=.8\linewidth]{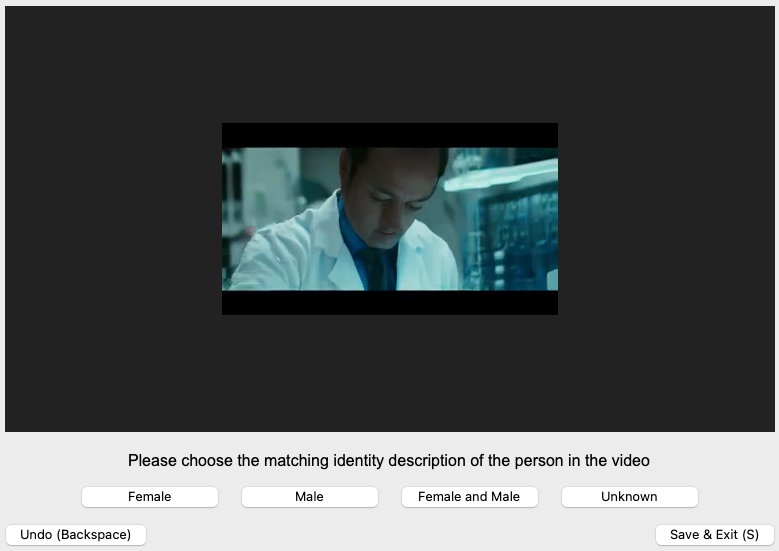}
    \caption{The screenshot of the user study GUI interface for the experiment on manual gender labelling (VFR-Human).}
    \label{fig:vfr-human}
\end{figure*}

We conduct the user study using structured survey forms with an interactive graphical user interface (GUI). Each human evaluation question is presented through a dedicated GUI module that displays the relevant generated video content together with a concise question description, which explicitly indicates the evaluation task and the available response options such as \Cref{fig:vfr-human}. Participants provide their responses by selecting from predefined choice buttons associated with each question.

Each evaluation prompt is presented as an independent section within the survey. For every section, participants are shown a single text prompt together with the corresponding generated videos and are asked to complete the associated evaluation task. Sections are isolated from one another, and no information from previous questions is visible, reducing potential carry-over effects and minimizing contextual bias between evaluations.

\begin{figure*}[ht]
    \centering
    \includegraphics[width=.8\linewidth]{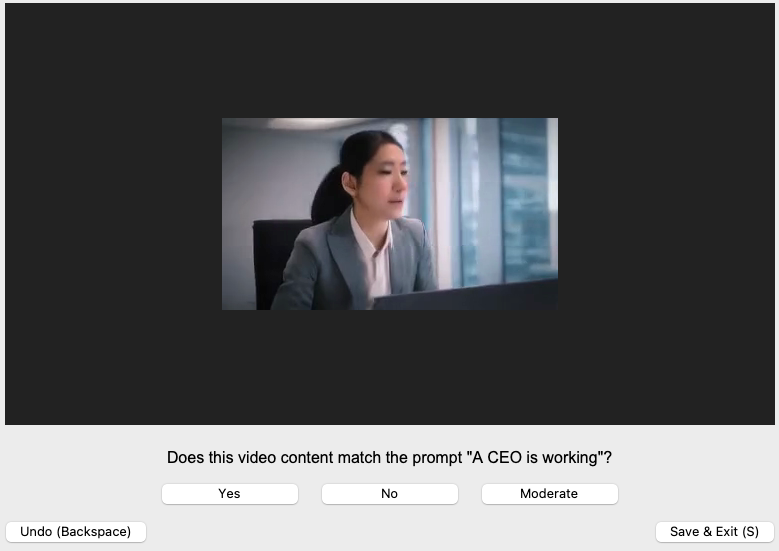}
    \caption{The screenshot of the user study GUI interface for the survey on human satisfaction
with generated videos.}
    \label{fig:human_match}
\end{figure*}

\begin{figure*}[ht]
    \centering
    \includegraphics[width=1.0\linewidth]{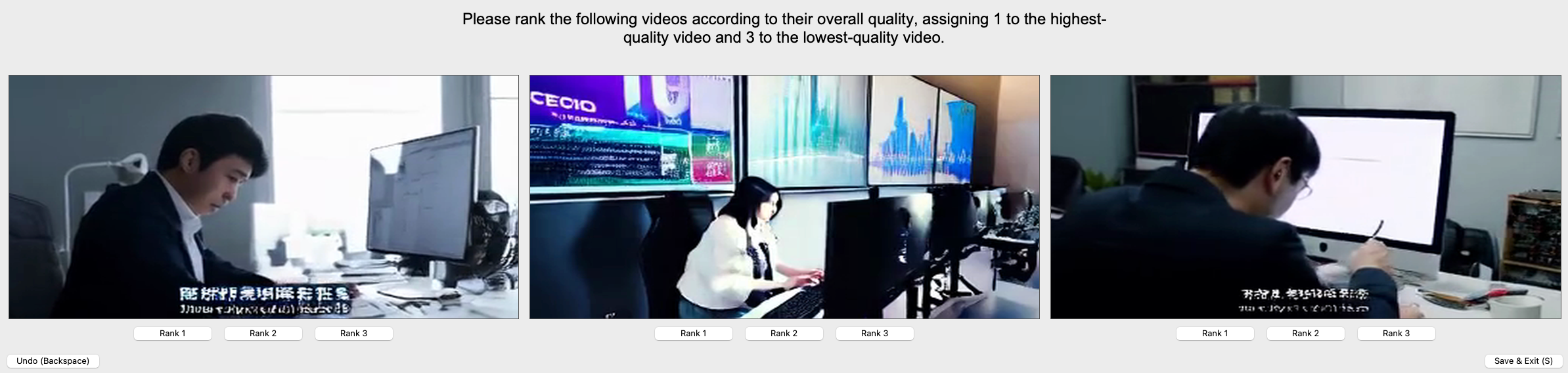}
    \caption{The screenshot of the user study GUI interface for the survey on human ranking of video quality.}
    \label{fig:human_rank}
\end{figure*}

In detail, as shown in \Cref{fig:vfr-human}, we use a dedicated interface for the manual identity labelling process used in the VFR-Human evaluation metric, where participants select the most appropriate identity description for the primary subject in the video. \Cref{fig:human_match} illustrates the GUI used for the text-video content alignment evaluation in~\Cref{sec:user-study}, and the corresponding quantitative results are reported in \Cref{fig:satisfaction}. For the other human ranking of video quality evaluation, the interface shown in \Cref{fig:human_rank} presents all videos generated from the same text prompt by different debiasing baselines and our method within a single section. The display order of videos is randomized for each evaluation instance to avoid position bias and to prevent method name leakage. The aggregated results of this ranking task are reported in \Cref{fig:rank}.

\subsection{Ethical Considerations}
The human participant study in this work was conducted after review and approval by the authors’ institutional ethics review board or was determined to be exempt under equivalent local procedures. Participants were recruited via an online platform and engaged in the study on a voluntary basis. All participants remained fully anonymous; no personally identifiable information was collected, stored, or shared. Prior to participation, individuals were informed about the nature of the study and provided consent according to the platform’s standard procedures.

To support reliable evaluation, participants were required to hold at least a bachelor’s degree or equivalent educational qualification. Responses were analyzed in aggregate, and individual contributions cannot be traced back to identifiable individuals. All data handling and reporting complied with applicable privacy regulations and research ethics practices relevant to human subjects research.